\documentclass[10pt,twocolumn,letterpaper]{article}

\usepackage{cvpr}
\usepackage{times}
\usepackage{epsfig}
\usepackage{graphicx}
\usepackage{amsmath}
\usepackage{amssymb}
\usepackage{color}
\usepackage{threeparttable}
\usepackage{enumitem}
\usepackage{authblk}

\usepackage[breaklinks=true,bookmarks=false]{hyperref}

\cvprfinalcopy 

\def\cvprPaperID{1369} 

\ifcvprfinal\fi
\begin{document}

\title{Transductive Unbiased Embedding for Zero-Shot Learning}
\author[1]{Jie Song}
\author[1]{Chengchao Shen}
\author[2]{Yezhou Yang}
\author[3]{Yang Liu}
\author[1]{Mingli Song}
\affil[1]{College of Computer Science and Technology, Zhejiang University, Hangzhou, China}
\affil[2]{Arizona State University, Tempe, USA}
\affil[3]{Alibaba Group, Hangzhou, China}

\maketitle
\thispagestyle{empty}

\begin{abstract}
Most existing Zero-Shot Learning (ZSL) methods have the strong bias problem, in which instances of unseen (target) classes tend to be categorized as one of the seen (source) classes. So they yield poor performance after being deployed in the generalized ZSL settings. In this paper, we propose a straightforward yet effective method named Quasi-Fully Supervised Learning (QFSL) to alleviate the bias problem. Our method follows the way of transductive learning, which assumes that both the labeled source images and unlabeled target images are available for training. In the semantic embedding space, the labeled source images are mapped to several fixed points specified by the source categories, and the unlabeled target images are forced to be mapped to other points specified by the target categories. Experiments conducted on AwA2, CUB and SUN datasets demonstrate that our method outperforms existing state-of-the-art approaches by a huge margin of $9.3\sim24.5\%$ following generalized ZSL settings, and by a large margin of $0.2\sim16.2\%$ following conventional ZSL settings.

\end{abstract}

\section{Introduction}
\label{section:introduction}
With the availability of large-scale training data, the field of visual object recognition has made significant progress in the last several years~\cite{krizhevsky2012imagenet,Simonyan14c,szegedy2015going,he2016deep,huang2017densely}. However, collecting and labeling training data are laboriously difficult and costly. For example, in fine-grained classification, expert knowledge is required to discriminate between different categories. For rare categories, such as endangered species, it's an extremely difficult work to collect sufficient and statistically diverse training images. Even worse, the frequencies of observing objects follow a long-tailed distribution~\cite{salakhutdinov2011learning,zhu2014capturing}, which indicates that the number of such unfrequent objects significantly surpasses that of common objects. Given limited or zero training images, existing visual recognition models (\eg, deep CNN models) struggle to make correct predictions.

\begin{figure}[t]
  \centering
  \includegraphics [scale=0.40]{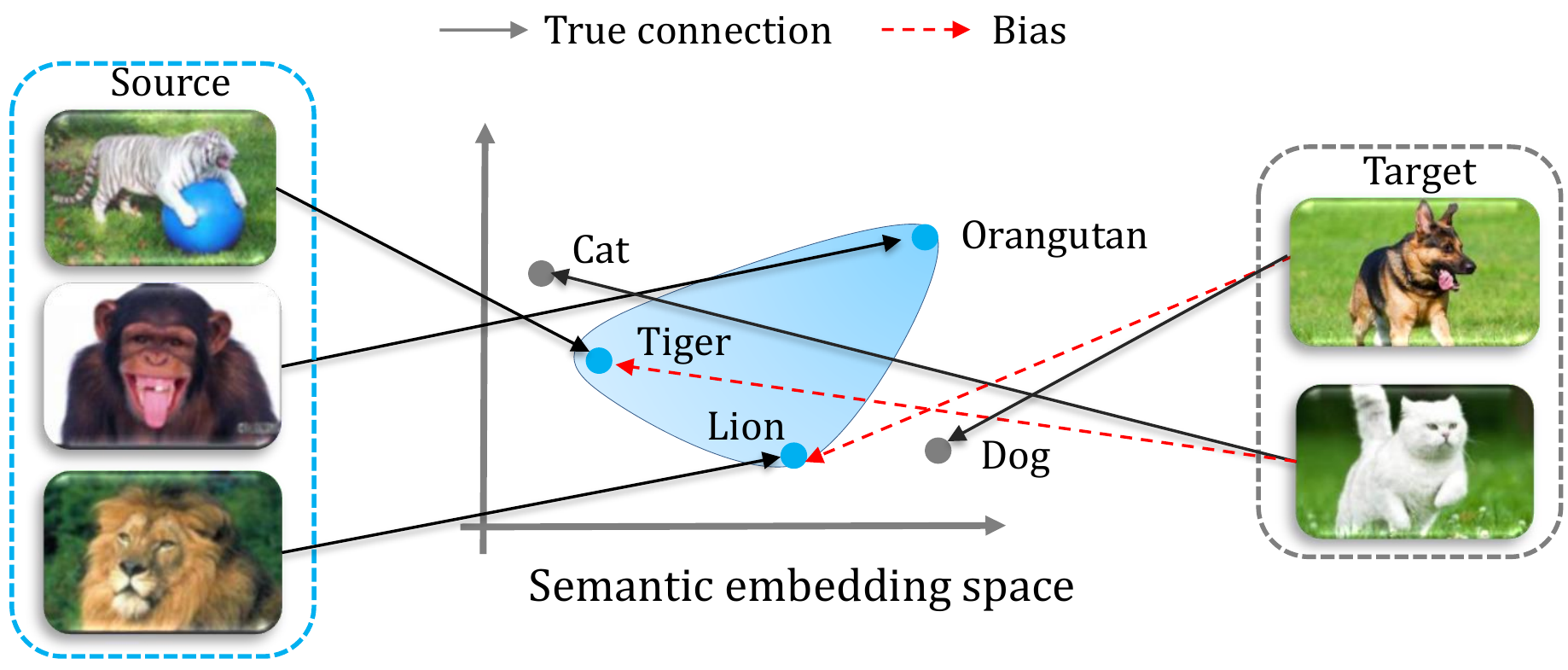}
  \caption{An illustrative diagram of the bias towards seen source classes in the semantic embedding space. The blue circles denote the anchor points specified by the source classes.}
  \label{fig:bias}
\end{figure}

Zero-Shot Learning (ZSL)~\cite{farhadi2009describing,lampert2009learning,akata2013label,romera2015embarrassingly,akata2015evaluation,reed2016learning,xian2017zero,MorgadoCVPR17} has emerged as a promising paradigm to alleviate the above problem. Unlike fully supervised classification which requires sufficient labeled training images for each category, ZSL distinguishes between two types of categories: \textit{source} and \textit{target}, where the labeled images are only available for the source categories. To facilitate the recognition of novel target categories, ZSL assumes the source and the target categories share a common semantic space to which both the images and class names can be projected. The semantic space can be defined by attributes~\cite{farhadi2009describing,akata2013label}, word2vec~\cite{mikolov2013efficient} or WordNet~\cite{miller1995wordnet}. Under this assumption, the recognition of images from novel target categories can be achieved by the nearest neighbor search in the shared space.

Depending on whether the unlabeled data of target classes are available for training, existing ZSL methods can be categorized into two schools: \textbf{inductive ZSL}~\cite{frome2013devise,norouzi2014zeroshot,zhang2015zero,akata2015evaluation,romera2015embarrassingly,changpinyo2016synthesized} and \textbf{transductive ZSL}~\cite{kodirov2015unsupervised,fu2015transductive,guo2016transductive}. For the inductive ZSL, only data of the source categories are available during the training phase. For the transductive ZSL methods, both the labeled source data and the unlabeled target data are available for training. The transductive ZSL aims to utilize the information from both the labeled source data and the unlabeled target data to accomplish the ZSL task.

During the test phase, most existing inductive and transductive ZSL methods~\cite{lampert2009learning,akata2013label,romera2015embarrassingly,akata2015evaluation,reed2016learning,MorgadoCVPR17} assume the test images come solely from the target classes. Therefore, the search space for classifying the new test images is restricted to the target classes. We call this experimental settings \textbf{conventional settings}. However, in a more practical situation, the test images come not only from the target but also from the source classes. Hence, both the source and the target classes should be considered. This experimental settings are usually regarded as the \textit{generalized} ZSL settings~\cite{xian2017zero,chao2016empirical}, abbreviated to \textbf{generalized settings} in this paper.

Existing ZSL methods perform much worse in the generalized settings than in the conventional settings~\cite{xian2017zero,chao2016empirical}. One vital factor accounting for the poor performance can be explained as follows. ZSL achieves the recognition of new categories by establishing the connection between the visual embeddings and the semantic embeddings. However, during the phase of bridging the visual and the semantic embeddings, there exists a strong bias~\cite{chao2016empirical} (shown in Figure~\ref{fig:bias}). During the training phase of most existing ZSL methods, the visual instances are usually projected to several fixed anchor points specified by the source classes in the semantic embedding space. This leads to a strong bias when these methods are used for testing: given images of novel classes in the target dataset, they tend to categorize them as one of the source classes.

To alleviate the mentioned problem above, we propose a novel transductive ZSL method in this paper. The proposed method assumes that both the labeled source and the unlabeled target data are available during the training phase. On the one hand, the labeled source data are used to learn the relationship between visual images and semantic embeddings. On the other hand, the unlabeled data of target classes are used to alleviate the strong bias towards source classes. More specifically, unlike other ZSL methods which always map input images to several fixed anchor points in the embedding space during training, our method allows the mapping from the inputs to other points, which significantly alleviates the strong bias problem.

\begin{figure*}
  \centering
  \includegraphics [scale=0.45]{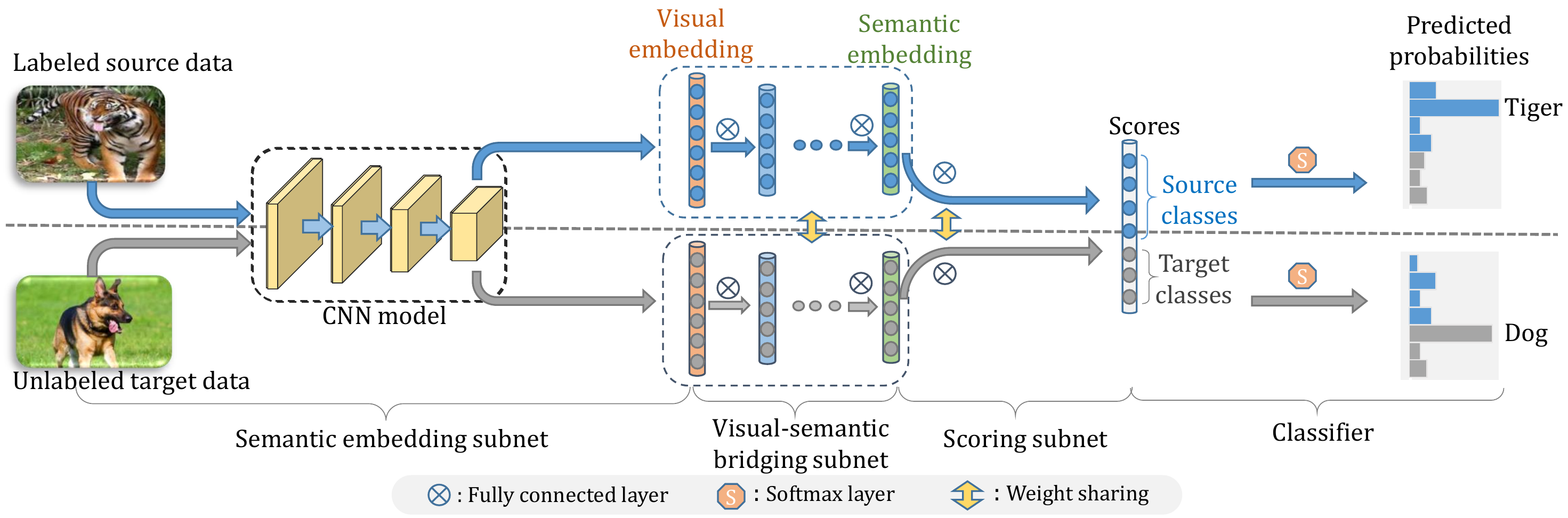}
  \caption{An overall architecture of the proposed QFSL model. Both the labeled and the unlabeled data are used to train the same model. Here for a better understanding, we depict them in two streams.}
  \label{fig:architecture}
\end{figure*}

We dub the proposed ZSL method as \textit{Quasi-Fully Supervised Learning} (QFSL), as it works like the conventional fully supervised classification in which a multi-layer neural network and a classifier are integrated together (shown in Figure~\ref{fig:architecture}). The architecture of the multi-layer neural network is usually taken from AlexNet~\cite{krizhevsky2012imagenet}, GoogleNet~\cite{szegedy2015going} or other well-known deep networks. In the training phase, our model is trained in an end-to-end manner to recognize the data from both source and target classes even without labeled data for the target classes. This feature brings up a compelling advantage: when the labeled data of target classes are available in the future, it can be directly used to train our model. In the test phase, our trained model can be directly used to recognize new images from both the source and the target classes without any modifications.

To sum up, we made the following contributions: 1) A transductive learning (QFSL) method is proposed to learn unbiased embeddings for ZSL. To our knowledge, this is the first work to adopt transductive learning method in solving the ZSL problem in generalized settings. 2) Experiments reveal that our method significantly outperforms existing ZSL methods, in both generalized and conventional settings.

\section{Related Work}
\label{section:related_work}
\paragraph{Zero-Shot Learning}
ZSL relies on the semantic space to associate source and target classes. Various semantic spaces have been investigated, including attributes~\cite{farhadi2009describing,lampert2009learning,akata2013label,xian2017zero,MorgadoCVPR17}, word vector~\cite{frome2013devise,miller1995wordnet}, text description~\cite{reed2016learning,zhang2016learning} and human gaze~\cite{karessliCVPR17}. The attribute has been shown to be an effective semantic space~\cite{akata2015evaluation,romera2015embarrassingly,MorgadoCVPR17} for ZSL. However, its superior performance is obtained at the cost of much more expensive human labor. As an alternative, the word vectors are gaining more attention recently~\cite{mikolov2013distributed,pennington2014glove} since they are learned from the large text corpus in an unsupervised way. Albeit their popularity, the word vectors often suffer from visual-semantic discrepancy problem~\cite{BMVC17Zeroshot,changpinyo2016predicting,demirel2017attributes2classname}. In addition to the word vectors, human gaze~\cite{karessliCVPR17} is recently proposed to replace the attributes, as its annotation can be performed by non-experts without domain knowledge.

In terms of the way how the visual space and the semantic space are related, existing ZSL methods can be mainly categorized into three groups: (1) from the visual space to the semantic space~\cite{frome2013devise,akata2015evaluation,reed2016learning}, (2) from the semantic space to the visual space~\cite{zhang2016learning,shigeto2015ridge,kodirov2015unsupervised} and (3) both the visual space and the semantic space are projected to a shared intermediate space~\cite{lu2016unsupervised,zhang2016zero,changpinyo2016synthesized}. As long as one of the above pathways is established, classification can be carried out via the nearest neighbor search in the embedding space which both the original visual inputs and the class labels can access. However, most existing ZSL methods share a common deficiency. During the training phase, regardless of how these two spaces are related, the existing ZSL usually project the visual inputs to several fixed points in the embedding space. It leads to the bias problem as discussed in Section~\ref{section:introduction}. Our work aims at alleviating this problem to improve the performance of ZSL.
\paragraph{Transductive Zero-Shot Learning}
Transductive ZSL solves ZSL in a semi-supervised learning manner where both the labeled source data and the unlabeled target data are available. Propagated Semantic Transfer (PST)~\cite{rohrbach2013transfer} exploits the manifold structure of novel classes to conduct label propagation. Transductive Multi-View ZSL (TMV)~\cite{fu2015transductive} and Unsupervised Domain Adaption (UDA)~\cite{kodirov2015unsupervised} associate cross-domain data by CCA and regularized sparse coding. In~\cite{guo2016transductive}, a joint learning approach is proposed to learn the Shared Model Space (SMS) for transductive ZSL settings. With the SMS, knowledge can be effectively transferred between classes using attributes. In this paper, we leverage both the labeled source data and the unlabeled target data to learn an unbiased embedding space for ZSL.
\paragraph{Zero-Shot Learning in Generalized Settings}
In performance evaluation, most existing ZSL methods usually assume that the test instances belong only to the unseen target classes. However, in practice, we are more often required to recognize instances from both the source and the target classes. The generalized settings relax the unrealistic assumption of the conventional settings with both the seen classes and the unseen classes at test time.  In~\cite{frome2013devise,norouzi2014zeroshot}, the source classes are considered when the classification is conducted, but only data from the unseen classes are tested. In~\cite{socher2013zero}, a two-stage approach is proposed to solve the ZSL problem in generalized settings. Before classification, it first determines whether a test instance is from a source or target class. In~\cite{chao2016empirical}, an empirical study and analysis of ZSL in generalized settings are provided. Recently, \cite{xian2017zero} shows many ZSL methods behave much worse in the generalized settings than in the conventional settings.

\section{Quasi-Fully Supervised Learning}
\label{section:QFSL}

\subsection{Problem Formulation}

Assume that there is a source dataset $\mathcal{D}^s = \{(x_i^s, y_i^s)\}_{i=1}^{N_s}$ consisting of $N_s$ images. Each image $x_i^s$ is associated with a corresponding label $y_i^s$, $y_i^s \in \mathcal{Y}^s=\{y_i\}_{i=1}^{S}$, and $S$ is the number of the source classes. Similarly, there is a target dataset $\mathcal{D}^t = \{(x_i^t, y_i^t)\}_{i=1}^{N_t}$ consisting of $N_t$ images. Each image $x_i^t$ is associated with a corresponding label $y_i^t$, $y_i^t \in \mathcal{Y}^t=\{y_{S+i}\}_{i=1}^{T}$, and $T$ is the number of the target classes. $\mathcal{Y}^s \cup \mathcal{Y}^t = \mathcal{Y}$, $\mathcal{Y}^s \cap \mathcal{Y}^t = \emptyset$. The goal of ZSL in conventional settings is to learn a prediction function $f$ as below from the source data

\begin{equation}
\label{eq:predict_function}
f(x; W) = \arg\max_{y\in \mathcal{Y}}F(x, y; W),
\end{equation}
so that its performance on the target data is maximized. $F$ is a score function, which ranks the correct label higher than the incorrect labels, and $W$ is the parameters of $F$. $F$ usually takes the following bilinear form~\cite{akata2013label,frome2013devise,akata2015evaluation}:
\begin{equation}
\label{eg:score_function}
F(x, y; W) = \theta(x)^TW\phi(y),
\end{equation}
where $\theta(x)$ and $\phi(y)$ are the visual and the semantic embeddings, respectively. The score function is usually optimized by minimizing the regularized loss:
\begin{equation}
L = \frac{1}{N_s}\sum_{i=1}^{N_s}L_p(y_i, f(x_i; W)) + \gamma\Omega(W),
\end{equation}
where $L_p$ is the classification loss (such as entropy loss and structured SVM~\cite{Tsochantaridis:2005:LMM:1046920.1088722}) to learn the mapping between the visual and the semantic embeddings. $\Omega$ is the regularization term used to constrain the complexity of the model.

In this paper, we assume the labeled source data $\mathcal{D}^s$, the unlabeled target data $\mathcal{D}_u^t=\{x_i^t\}_{i=1}^{N_t}$, and the semantic embeddings $\phi$ are available for training in our approach. The aim of our method is to achieve good performance in not only the conventional but also the generalized settings.

\subsection{The QFSL Model}

Different from the bilinear form described above, the scoring function $F$ in our method is designed as a nonlinear one. The whole model is implemented by a deep neural network (shown in Figure~\ref{fig:architecture}). It consists of four modules: the visual embedding subnet, the visual-semantic bridging subnet, the scoring subnet, and the classifier. The visual embedding subnet maps the raw images into visual embedding space. The visual-semantic bridging subnet projects the visual embeddings to semantic embeddings. The scoring subnet produces scores of every class in the semantic embedding space. And the classifier makes the final predictions based on the scores. All modules are differentiable and implemented by widely used layers including the convolutional layer, the fully connected layer, the ReLU~\cite{krizhevsky2012imagenet} layer and the softmax layer. Hence, our model can be trained in an end-to-end manner. Now we describe each module in detail in the following sections.

\subsubsection{Visual Embedding Subnet}

Most existing ZSL models~\cite{Fu_2016_CVPR,akata2015evaluation,DBLP:conf/eccv/BucherHJ16,romera2015embarrassingly,zhang2015zero,lampert2014attribute} adopt deep CNN features for visual embeddings. The visual embedding function $\theta$ is fixed in these methods. So they do not fully exploit the power of deep CNN models. Here, we also adopt a pre-trained CNN model to perform visual embedding. The major difference is that our visual embedding function can be optimized together with other modules\footnote{In some situations, keeping the visual embedding subnet fixed produces better performance. We conduct further discussions in Section~\ref{section:fixed_or_not}.}. The parameters of the visual embedding subnet are denoted by $W_{\theta}$. Unless otherwise specified, we use the output of the first fully connected layer as the visual embeddings.

\subsubsection{Visual-Semantic Bridging Subnet}

It is vital to build the connections between the image and the semantic embeddings. The connection can be built by either a linear~\cite{akata2013label,frome2013devise,akata2015evaluation} or a nonlinear~\cite{7780384,socher2013zero} function. In this paper, we adopt a non-linear function $\varphi$ to project the visual embeddings to the semantic embeddings. $\varphi$ is implemented by several fully connected layers, each of which is followed by a ReLU non-linear activation layer. The design of bridging function depends on the CNN architecture from the visual embedding subnet. Specifically, our design follows the fully connected layers of the selected CNN model. The visual-semantic bridging subnet is optimized together with the visual embedding subnet. The parameters of the visual-semantic bridging subnet are denoted by $W_{\varphi}$.

\subsubsection{Scoring Subnet}

After bridging the visual and the semantic embeddings, recognition task can be carried out by the nearest neighbor search in the semantic embedding space. Given an image, we firstly obtain its visual embedding by the visual embedding subnet. Then the visual embedding is mapped to the semantic embedding by the visual-semantic bridging subnet. Finally, we use the inner product between the projected embedding and the normalized semantic embeddings as the scores. Therefore, the score function is

\begin{equation}
\label{eg:our_score_function}
F(x, y; W) = \varphi(\theta(x; W_{\theta}); W_{\varphi})\phi^{*}(y)
\end{equation}
where $W_{\theta}$ and $W_{\varphi}$ are the weights of the visual embedding function and the visual-semantic bridging function respectively, and $\phi^{*}(y)$ is the normalized semantic embedding of $y$: $\phi^{*}(y) = \frac{\phi(y)}{\left\|\phi(y)\right\|_{2}}$.

The scoring subnet is implemented as a single fully connected layer. The weights are initialized with the normalized semantic vectors of both the source and the target classes: $[\phi^{*}(y_1), \phi^{*}(y_2), ..., \phi^{*}(y_{S+T})]$. Unlike the visual embedding subnet and the visual-semantic bridging subnet, the weights of the scoring subnet are frozen and will not be updated during the training phase. In this way, for a labeled source image ($x_i^s$, $y_i^s$), our model is trained to project the image $x_i^s$ to an embedding which has the most similar direction with the semantic embedding $\phi(y_i^s)$.

Note that though we don't have the labeled data of target classes, the target classes will also be involved in the training in our approach.  Hence during the training phase, our method produces $S+T$ scores for a given image.

\subsubsection{Classifier}

After the scoring subnet, we apply a traditional ($S+T$)-way softmax classifier to produce the predicted probability vector for all the classes. The predicted class of the input image is just the one with the highest probability.

\subsection{Optimization of the QFSL Model}
\label{section:optimization}
As described above, the architecture of our method is like the conventional fully supervised classification model with a ($S+T$)-way classifier for both the target and the source classes. Unfortunately, only the data for source classes are labeled while the data from target classes is unlabeled. In order to train the proposed model, we define a  \textit{Quasi-Fully Supervised Learning (QFSL)} loss:
\begin{equation}
\begin{split}
L = & \frac{1}{N_s}\sum_{i=1}^{N_s}L_p(x_i^s) + \frac{1}{N_t}\sum_{i=1}^{N_t}\lambda L_b(x_i^t) + \gamma\Omega(W).
\end{split}
\end{equation}
It is known that the  loss of conventional fully supervised classification is usually composed by the classification loss $L_p$ and regulation loss $\Omega$. Different from such conventional definition, our proposed QFSL incorporates an additional bias loss $L_b$ to alleviate the bias towards source classes:
\begin{equation}
L_b(x_i^t)= -\ln\sum_{i\in\mathcal{Y}^t}p_i,
\end{equation}
where $p_i$ is the predicted probability of class $i$. Given unlabeled instances from the target classes, this loss encourages our model to increase the sum of probabilities of being any target class. And consequently the model will prevent the instances of target classes from being mapped to the source classes.

For the classification loss $L_p$, we adopt the entropy loss in our method. For the regularization loss $\Omega$, $\ell^2$-norm is used for all the trainable parameters $W = \{W_\theta, W_\varphi\}$. $\lambda$ and $\gamma$ are trade-off weights among different losses, and they are set via cross-validation.

During the training phase, all the labeled and unlabeled data are mixed for training. Our model is optimized by the stochastic gradient descent algorithm. Each batch of training images is randomly drawn from the mixed dataset. Although our method is straightforward without bells and whistles, experiments show that it not only significantly alleviates the bias problem but also facilitates the building of connections between visual and semantic embeddings.
\section{Experiments}
\label{section:experiments}
In this section, extensive experiments are carried out to evaluate the performance of the proposed QFSL method. Firstly, we introduce some basic experimental settings. Then we discuss two implementation details of our method. Finally, we compare our proposed QFSL with existing state-of-the-art ZSL methods, in both the conventional and the generalized settings.
\subsection{Experimental Settings}
\paragraph{Datasets}
Three datasets are considered: Animals with Attributes 2 (AwA2)~\cite{xian2017zero}, Caltech-UCSD Birds-200-2011 (CUB)~\cite{WelinderEtal2010} and SUN Attribute Database (SUN)~\cite{Patterson2012SunAttributes}. AwA2 is a coarse-grained dataset. It contains 37,322 images of 50 animals classes, in which 40 classes are used for training and the rest 10 classes for testing. For each class, there are about 750 labeled images. CUB is a fine-grained dataset containing 11,788 images of 200 bird species. We use 150 classes for training and the rest 50 for testing. In this dataset, each class has about 60 labeled images. SUN is another fine-grained dataset. There are 14,340 images coming from 717 types of scenes, of which 645 types are used for training, and the rest 72 for testing. Note that there are only about 20 images for every class on SUN, which is relatively scarce. In our experiments, we adopt either the standard train/test splits (SS) or the splits proposed (PS) in~\cite{xian2017zero} in some experiments for fair comparisons.

Class-level attributes are used in our experiments. For AwA2, we use the provided continuous 85-dimension class-level attributes~\cite{xian2017zero}. For CUB, continuous 312-dimension class-level attributes are provided in~\cite{WelinderEtal2010}. For SUN, there are continuous 102-dimension attributes provided in~\cite{Patterson2012SunAttributes}.
\paragraph{Model Selection and Training}
Four popularly used deep CNN models are involved in our following experiments: AlexNet~\cite{krizhevsky2012imagenet}, GoogLeNet~\cite{szegedy2015going}, VGG19~\cite{Simonyan14c} and ResNet101~\cite{xian2017zero}. They are all pre-trained on ImageNet~\cite{ILSVRC15} with 1K classes. Among these models, GoogLeNet is one of the most popular models used in the ZSL field, so we adopt GoogLeNet when we make comparisons between our and existing methods.

Unless otherwise specified, the learning rate is fixed to be 0.001, and the minibatch size is 64. The scaling weights of bias loss ($\lambda$) and weights decay ($\gamma$) are 1 and 0.0005, respectively. The training process stops after 5,000 iterations. These hyper-parameters are selected based on class-wise cross validation~\cite{zhang2015zero,changpinyo2016synthesized,chao2016empirical}.

\paragraph{Evaluation Metrics}
\label{section:performance_metric}
To compare the performances, we adopt the Mean Class Accuracy (MCA) as the evaluation metric in our experiments:
\begin{equation}
\label{eq:mca}
MCA = \frac{1}{|\mathcal{Y}|}\sum_{y\in \mathcal{Y}} acc_y,
\end{equation}
where $acc_y$ is the top-1 accuracy on the test data from class $y$. In the conventional settings, MCA on only the target test data ($MCA_t$) is considered ($\mathcal{Y}=\mathcal{Y}_t$ in Eqn.~\ref{eq:mca}). In the generalized settings, the search space at evaluation time is not restricted to the target classes, instead the the source classes are also included. Meanwhile, the test instances come from not only the target dataset, but also the source
dataset ($\mathcal{Y}=\mathcal{Y}_s+\mathcal{Y}_t$ in Eqn.~\ref{eq:mca}). Therefore, we adopt $MCA_t$, $MCA_s$ (MCA on the source test data) and their harmonic mean ($H$) as the evaluation metrics:

\begin{equation}
H = \frac{2*MCA_s*MCA_t}{MCA_s + MCA_t}.
\end{equation}

\subsection{Implementation Discussions}
\label{section:discussion}

\subsubsection{Optimization of the Visual Embedding Subnet}
\label{section:fixed_or_not}
\begin{figure}
  \centering
  \includegraphics [scale=0.32]{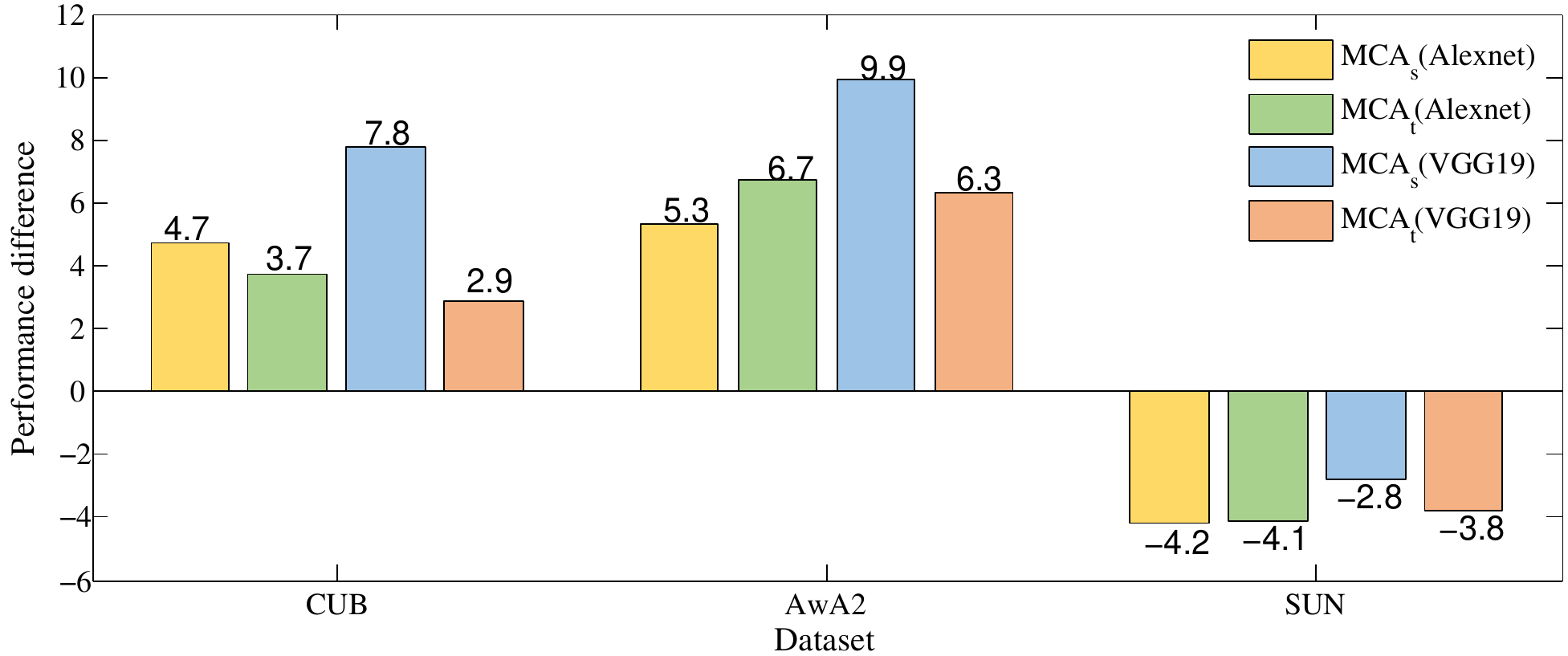}
  \caption{Comparisons between optimizing the visual embedding subnet and keeping it fixed. Performance difference = $MCA$(unfixed) $-$ $MCA$(fixed).}
  \label{fig:image_embedding_fixed_or_not}
\end{figure}
\begin{figure}
  \centering
  \includegraphics [scale=0.32]{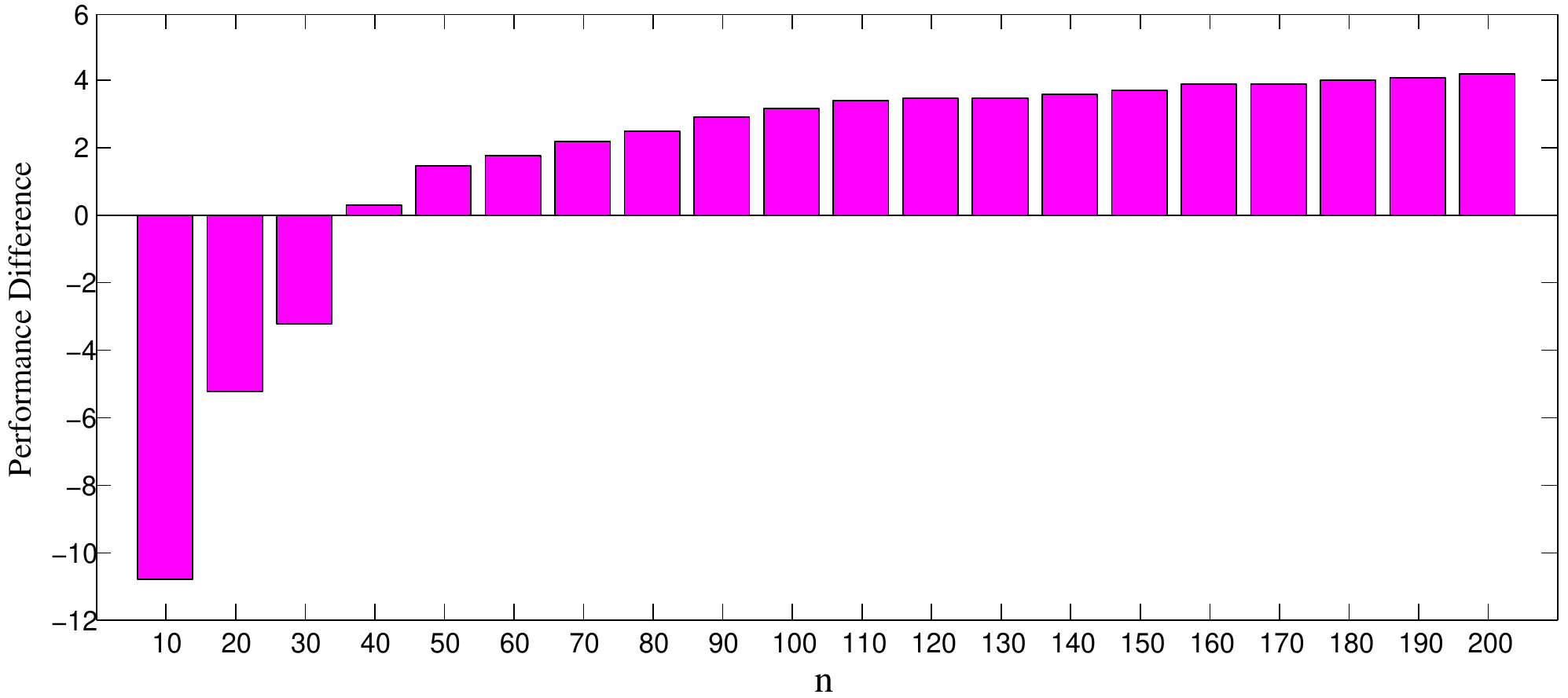}
  \caption{Performance difference with different number of training images per class on AwA2.}
  \label{fig:varying_images_per_class}
\end{figure}
Many existing ZSL methods adopt pre-trained deep ConvNets as the visual embedding function. Most of them keep the trained CNN models fixed and do not optimize them during the training phase. In contrast, in our method, the visual embedding subnet can be optimized together with other parts. In this experiment, we compare the performance of our method between with and without the visual embedding subnet fixed. All four models (AlexNet, GoogLeNet, VGG19, and ResNet101) are adopted to implement our methods. Experiments in the generalized settings are conducted on CUB, AwA2 and SUN datasets. The results of AlexNet and VGG19 are shown in Figure~\ref{fig:image_embedding_fixed_or_not} (GoogLeNet and ResNet101 produce similar results). It can be seen that with the visual embedding subnet optimized, QFSL achieves much better performance on CUB and AwA2 than that with visual embedding function fixed. However, on the SUN dataset, training the visual embedding subnet produces a worse performance. We speculate that the scarce training data for source classes account for that. On AwA2 and CUB, there are about 750 and 60 training images for each category, respectively. However, there are only 20 images for each category on SUN. To validate our speculation, we conduct another experiment on the AwA2 dataset, as there are much more images per class in this dataset. In this experiment, our model is trained with different numbers (denoted by $n$) of labeled source images per class. Results are depicted in Figure~\ref{fig:varying_images_per_class}. It can be concluded that with fewer training images per class, training the visual embedding subnet indeed leads to worse performance, which verifies our speculation.

\subsubsection{Classification Loss and Bias Loss}
\begin{figure}
  \centering
  \includegraphics [scale=0.32]{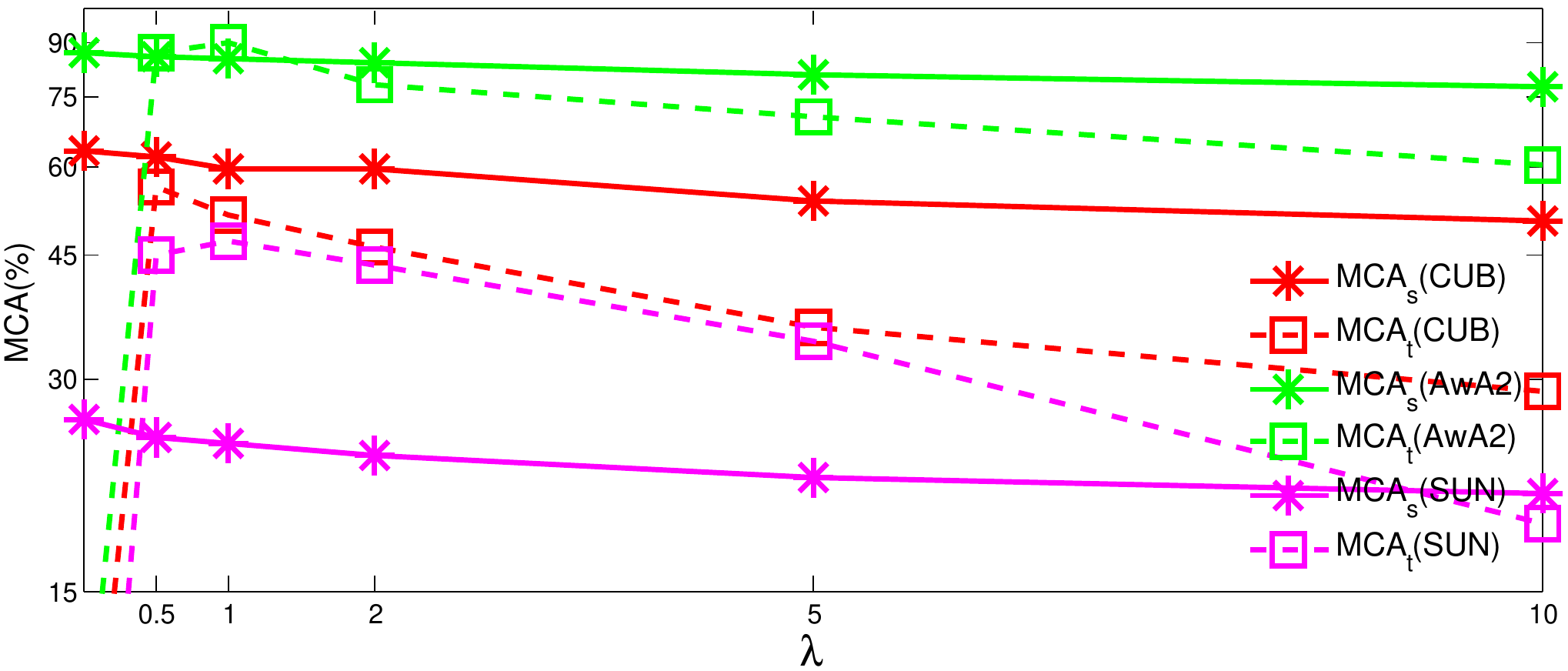}
  \caption{Performance of QFSL with varying $\lambda$.}
  \label{fig:trade-off}
\end{figure}
As aforementioned in Section~\ref{section:optimization}, there are three components in our loss function: the classification loss, the bias loss, and the regularization loss. The classification loss is used to build the connection between the visual embeddings and the semantic embeddings, and the bias loss is designed to alleviate the bias towards source classes. In this section, we explore how the trade-off between the classification loss and the bias loss impacts the performance of QFSL in the generalized settings.

We test QFSL with several different $\lambda$ values $\{0.0, 0.5, 1.0, 2.0, 5.0, 10.0\}$ on all the three datasets. In the experiment, we adopt the AlexNet as the visual embedding function. Figure~\ref{fig:trade-off} shows the results of QFSL with different $\lambda$. Consistently, on all the three datasets, $MCA_s$ decreases steadily as we increase $\lambda$. It is reasonable because putting more attention to alleviating the bias will distract the model from building the connection between image and semantic embeddings. For $MCA_t$, the overall best results are obtained when $\lambda \in [0.5, 2]$. Smaller $\lambda$ ($< 0.5$) leaves the bias problem unsolved. On the other side, larger $\lambda$ ($> 2$) yields negative effects on the building of the relationship between image and semantic embeddings, thus decreasing $MCA_t$ in return.

\subsection{Comparisons in Conventional Settings}
\label{section:comparison_with_inductive}
\begin{table}
\scriptsize
\begin{center}
\caption{Comparisons in conventional settings (in \%). For each dataset, the best result is marked in \textbf{bold} font and the second best in blue. We report results averaged over 5 random trails.}
\label{table:inductive_zsl}
\begin{threeparttable}
\begin{tabular}{c|c|cc|cc|cc}
\hline
  \multicolumn{1}{c|}{} & \multicolumn{1}{c|}{}  & \multicolumn{2}{c|}{\textbf{CUB}}    & \multicolumn{2}{c|}{\textbf{SUN}}     & \multicolumn{2}{c}{\textbf{AwA2}}  \\

&\textbf{Method} & \textbf{SS} & \textbf{PS} & \textbf{SS} & \textbf{PS} & \textbf{SS} &\textbf{PS}\\ \hline
&DAP~\cite{lampert2014attribute} & 37.5 & 40.0 & 38.9 & 39.9 & 58.7 & 46.1\\
&CONSE~\cite{norouzi2014zeroshot} & 36.7 & 34.3 & 44.2  & 38.8 & 67.9 & 44.5\\
&SSE~\cite{zhang2015zero} & 43.7 & 43.9 & 25.4 & 54.5 & 67.5 & 61.0\\
&ALE~\cite{akata2013label} & 53.2 & 54.9 & 59.1 & {\color{blue}58.1} & {\color{blue}80.3} & 62.5\\
\S&DEVISE~\cite{frome2013devise} & 53.2 & 52.0 & 57.5 & 56.5 & 68.6 & 59.7\\
&SJE~\cite{akata2015evaluation} & 55.3 & 53.9 & 57.1 & 53.7 & 69.5 & 61.9\\
&ESZSL~\cite{romera2015embarrassingly} & 55.1 & 53.9 & 57.3 & 54.5 & 75.6 & 58.6\\
&SYNC~\cite{changpinyo2016synthesized} & 54.1 & 55.6 & 59.1 & 56.3 & 71.2 & 46.6\\
\hline
&UDA~\cite{kodirov2015unsupervised} & 39.5 & -- & -- & -- & -- & --\\
\pounds&TMV~\cite{fu2015transductive} & 51.2 & -- & {\color{blue}61.4} & -- & -- & --\\
&SMS~\cite{guo2016transductive} & {\color{blue}59.2} & -- & 60.5 & -- & -- & --\\
\hline
&QFSL$^-$ & 58.5 & {\color{blue}58.8} & 58.9 & 56.2 & 72.6 & {\color{blue}63.5}\\
& &$_{\color{green}\uparrow10.5}$ &$_{\color{green}\uparrow13.3}$ & $_{\color{green}\uparrow0.3}$& $_{\color{green}\uparrow0.2}$& $_{\color{green}\uparrow4.5}$ &$_{\color{green}\uparrow16.2}$\\
&QFSL & \textbf{{69.7}} & \textbf{{72.1}} & \textbf{{61.7}} & \textbf{{58.3}} & \textbf{{84.8}}  &\textbf{{79.7}}\\
\hline
\end{tabular}
\begin{tablenotes}
   \item [\S]: inductive ZSL methods.
   \item [\pounds]: transductive ZSL methods.
   \item [${\color{green}\uparrow}$]: performance boost compared with the best existing ZSL methods (including the baseline QFSL$^-$).
\end{tablenotes}
\end{threeparttable}
\end{center}
\end{table}
\begin{figure}
  \centering
  \includegraphics [scale=0.35]{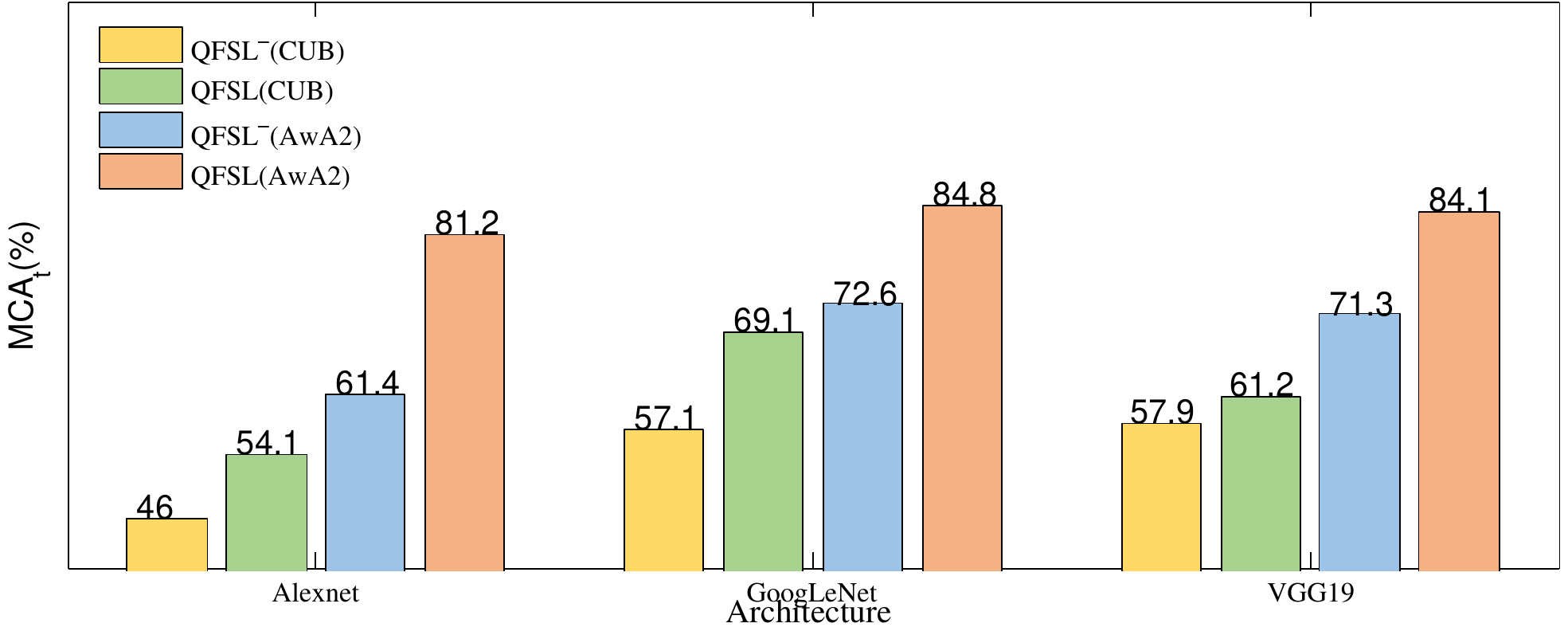}
  \caption{Comparisons between QFSL$^-$ and QFSL on different CNN architectures.}
  \label{fig:multi-models}
\end{figure}

\begin{table*}
\scriptsize
\begin{center}
\caption{Comparisons in the generalized settings (in \%). Previously published results are given in normal font, and results of our implementations are given in \textit{italics} font. For\textit{ QFSL$^G$} and \textit{QFSL$^R$}, the visual embedding function is implemented with GoogLeNet and ResNet101, respectively. For each dataset, the best result is marked in \textbf{bold} font and the second best in blue. We report results averaged over 5 random trails (CMT$^*$: CMT with novelty detection).}
\label{table:generalized_zsl}
\begin{threeparttable}
\begin{tabular}{c|c|ccc|ccc|ccc}
\hline
  \multicolumn{1}{c|}{}  &\multicolumn{1}{c|}{} & \multicolumn{3}{c|}{\textbf{AwA2}}    & \multicolumn{3}{c|}{\textbf{CUB}}     & \multicolumn{3}{c}{\textbf{SUN}}  \\
&\textbf{Method} & \textbf{MCA}$_s$ & \textbf{MCA}$_t$ & \textbf{H} & \textbf{MCA}$_s$ & \textbf{MCA}$_t$ & \textbf{H} & \textbf{MCA}$_s$ & \textbf{MCA}$_t$ & \textbf{H} \\ \hline
&DAP~\cite{lampert2014attribute} & 84.7 & 0.0 & 0.0  & 67.9 & 1.7 & 3.3 & 25.1 & 4.2 & 7.2\\
&CONSE~\cite{norouzi2014zeroshot} & 90.6 & 0.5 & 1.0 & 72.2 & 1.6 & 3.1 & \textbf{39.9} & 6.8 & 11.6\\
&SSE~\cite{zhang2015zero} & 82.5 & 8.1 & 14.8 & 46.9 & 8.5 & 14.4 & {\color{blue}36.4} & 2.1 & 4.0\\
\dag&ALE~\cite{akata2013label} & 81.8 & 14.0 & 23.9 & 62.8 & 23.7 & 34.4 & 33.1 & 21.8 & 26.3\\
&DEVISE~\cite{frome2013devise} & 74.7 & 17.1 & 27.8 & 53.0 & 23.8 & 32.8 & 30.5 & 14.7 & 19.8\\
&SJE~\cite{akata2015evaluation} & 73.9 & 8.0 & 14.4 & 59.2 & 23.5 & 33.6 & 30.5 & 14.7 & 19.8\\
&ESZSL~\cite{romera2015embarrassingly} & 77.8 & 5.9 & 11.0 & 63.8 & 12.6 & 21.0 & 27.9 & 11.0 & 15.8\\
&SYNC~\cite{changpinyo2016synthesized}  & 90.5 & 10.0 & 18.0 & 70.9 & 11.5 & 19.8 & 43.3 & 7.9 & 13.4\\
&CMT~\cite{socher2013zero} & 90.0 & 0.5 &1.0 &49.8  & 7.2 &12.6 &21.8  & 8.1 & 11.8\\
\hline
&CMT$^*$~\cite{socher2013zero} & 89.0 & 8.7 &15.9 &60.1  & 4.7 &8.7 &28.0  & 8.7 & 13.3\\
\ddag&\textit{CS}~\cite{chao2016empirical} & \textit{77.6} & \textit{45.3} & \textit{57.2} & \textit{49.4} & \textit{48.1} & \textit{48.7} & \textit{22.0} & \textit{44.9} & \textit{29.5}\\
&\textit{baseline} & \textit{72.8} & \textit{52.1} &\textit{60.7} &\textit{48.1}  & \textit{33.3} &\textit{39.4} &\textit{18.5} & \textit{30.9} & \textit{23.1}\\
\hline
&\textit{QFSL$^G$} & \textit{{\color{blue}92.4}$^{\color{green}{\uparrow1.8}}$} & \textit{{\color{blue}64.3}$^{\color{green}{\uparrow12.2}}$} & \textit{{\color{blue}75.8}$^{\color{green}{\uparrow15.1}}$} & \textit{{\color{blue}74.2}$^{\color{green}{\uparrow2.0}}$} & \textbf{\textit{{71.6}$^{\color{green}{\uparrow23.5}}$}} & \textit{{\color{blue}72.9}$^{\color{green}{\uparrow24.2}}$} & \textit{33.6}$^{\color{red}{\downarrow6.3}}$ & \textbf{\textit{{54.8}$^{\color{green}{\uparrow9.9}}$}} & \textbf{\textit{{41.7}$^{\color{green}{\uparrow12.2}}$}}\\
\ddag&\textit{QFSL$^R$} & \textbf{\textit{{93.1}$^{\color{green}{\uparrow2.5}}$}} & \textbf{\textit{{66.2}$^{\color{green}{\uparrow14.1}}$}} & \textbf{\textit{{77.4}$^{\color{green}{\uparrow16.7}}$}} & \textbf{\textit{{74.9}$^{\color{green}{\uparrow2.7}}$}} & \textit{{\color{blue}71.5}$^{\color{green}{\uparrow23.4}}$} &\textbf{\textit{{73.2}$^{\color{green}{\uparrow24.5}}$}} & \textit{31.2$^{\color{red}{\downarrow8.7}}$} & \textit{{\color{blue}51.3}$^{\color{green}{\uparrow6.4}}$} & \textit{{\color{blue}38.8}$^{\color{green}{\uparrow9.3}}$}\\
\hline
\end{tabular}
\begin{tablenotes}
   \item [$\dag$]: ZSL methods which \textit{do not} takes generalized settings into consideration.
   \item [$\ddag$]: ZSL methods which takes generalized settings into consideration.
   \item [${\color{green}\uparrow}$]: Performance boost compared with the best existing ZSL methods (including the baseline).
   \item [${\color{red}\downarrow}$]: Performance drop compared with the best existing ZSL methods (including the baseline).
\end{tablenotes}
\end{threeparttable}
\end{center}
\end{table*}

We firstly compare our method with existing state-of-the-art ZSL methods in the conventional settings. The compared methods include: 1) \textbf{inductive methods} DAP~\cite{lampert2014attribute}, CONSE~\cite{norouzi2014zeroshot}, SSE~\cite{zhang2015zero}, ALE~\cite{akata2013label}, DEVISE~\cite{frome2013devise}, SJE~\cite{akata2015evaluation}, ESZSL~\cite{romera2015embarrassingly}, SYNC~\cite{changpinyo2016synthesized}, and 2) \textbf{transductive methods} UDA~\cite{kodirov2015unsupervised}, TMV~\cite{fu2015transductive} and SMS~\cite{guo2016transductive}. In addition to these existing ZSL methods, there exists a latent baseline: training our proposed model with only labeled source data, \ie, the inductive version of our model. In this case, QFSL loss degrades to conventional fully supervised classification loss. We denote this baseline by QFSL$^-$ and also compare our method with it.

Experiments are conducted on AwA2, CUB, and SUN. We use both the standard split (SS) and the proposed split (PS)~\cite{xian2017zero} for more convincing results. The visual embedding subnet is optimized for AwA2 and CUB, but fixed for SUN. Table~\ref{table:inductive_zsl} shows the experimental results. It can be seen that 1) the baseline of our method (QFSL$^-$) yields comparable performance with existing ZSL methods, and 2) the proposed method outperforms the baseline and existing approaches on all datasets. Notably, on CUB and AwA2, our method outperforms other state-of-the-art ZSL methods (including QFSL$^{-}$) by a large margin of $4.5\sim16.2\%$. The experimental results indicate that our approach effectively utilizes the valuable information contained in the unlabeled target data to facilitate the building of connections between the visual and the semantic embeddings.


To further verify that our method is not only effective to a specific CNN model, we implement our method with AlexNet, GoogleNet, and VGG respectively. In this experiment, as QFSL$^-$ is shown to achieve comparable performance with other ZSL methods in Table~\ref{table:inductive_zsl}, we compare our method only with QFSL$^-$. The comparison result is provided in Figure~\ref{fig:multi-models}. It can be noticed that our method outperforms the baseline consistently on all the three CNN models, which validates the effectiveness of our method.

\subsection{Comparisons in Generalized Settings}
\label{section:comparison_generalized_zsl}

Our method is designed to alleviate the strong bias problem. Therefore, we verify its effectiveness in the generalized settings, in which the strong bias problem often leads to poor performance. Before evaluating the performance of our method, there remains one issue to address. When evaluating the performance in the test phase, most of the existing transductive ZSL methods use the same target data used in the training phase. However, if our method adopts the same policy, it will be problematic because our method has already used the supervisory information that the unlabeled data are coming from the target classes. To solve this problem, we split the unlabeled target data into two halves and train two QFSL models. One half of the unlabeled data is used for training and the other one for testing when training our first model, and vice versa when training our second model. The final performance of our method is the average performance of these two models. To our knowledge, this is the first study on applying the transductive method to solve the ZSL problem in generalized settings.

We compare our method with several state-of-the-art ZSL methods~\cite{lampert2014attribute,norouzi2014zeroshot,zhang2015zero,akata2013label,frome2013devise,akata2015evaluation,romera2015embarrassingly,changpinyo2016synthesized}. However, these methods do not take the generalized settings into consideration. In addition to these methods, we also compare our methods with two other ZSL methods \textit{Calibrated Stacking (CS)}~\cite{chao2016empirical} and \textit{Cross Model Transfer (CMT)}~\cite{socher2013zero}, which take the generalized settings into consideration. CS maximizes the performance in the generalized settings by trading off between $MCA_s$ and $MCA_t$. CMT first utilizes novelty detection methods~\cite{socher2013zero} to differentiate between source and target classes and then accordingly applies the corresponding classifiers. As our method utilizes the unlabeled target data, we introduce another baseline (called \textit{baseline} here) for a clearer comparison. The \textit{baseline} trains a deep binary classifier (GoogLeNet) on the available source data and unlabeled target data to discriminate between the source and the target data, then classifies the test instances in the corresponding search space.

The original data split and other experimental settings are kept the same as that used in~\cite{xian2017zero}, where the visual embedding function is implemented with ResNet101. For a fair comparison, we also adopt ResNet101 to implement the visual embedding function (denoted by QFSL$^R$). In addition, as GoogLeNet is widely used in ZSL, the performance of our method with GoogLeNet is also provided (denoted by QFSL$^G$). Experimental results are given in Table~\ref{table:generalized_zsl}. It can be seen that generally our method improves the overall performance (harmonic mean $H$) by an obvious margin ($9.3\sim24.5\%$ on the three datasets). The performance boost mainly comes from the improvement of mean class accuracy on the target classes ($MCA_t$), meanwhile without much performance drop on the source classes ($MCA_s$). These compelling results verify that our method can significantly alleviate the strong bias towards source classes by using the unlabeled instances from the target classes.

Another noticeable result from Table~\ref{table:generalized_zsl} is that the results of QFSL$^R$ are generally better that of QFSL$^G$ on CUB and AwA2 datasets. However, on SUN, QFSL$^G$ achieves better performance. We observe the fact that only scarce (about 20) training images are available for each source category in the SUN dataset accounts for that. Using such scarce data to train deep CNN models like ResNet101 usually leads to over-fitted models. 

\section{Further Study and Discussions}
\label{section:imbalance}
In real-world scenarios, the number of the target classes usually greatly surpasses that of source ones. However, most datasets for ZSL benchmark violate that. For examples, for AwA2, only 10 of 50 classes are treated as target ones. On CUB, only 50 out of 150 classes are used as the target. More severely, on the SUN dataset, only 72 out of 717 classes are put into the target classes. In this section, we empirically study how the imbalance between the source and the target classes affects the proposed QFSL method.

Experiments are conducted on the SUN dataset, as there are much more classes in it. The visual embedding function is implemented with GoogLeNet. We adopt the standard split used by the most of other works. 72 classes are treated as the target categories. For the source categories, we randomly select seven subsets from the rest categories. The number of source categories is $\{100, 200, 300, 450, 550, 600, 645\}$. We use these 7 different source data and the fixed target data to test out method. For a better understanding of our method, we also depict the performance of the baseline QFSL$^-$, in which only the labeled source data are available.

\begin{figure}
  \centering
  \includegraphics [scale=0.35]{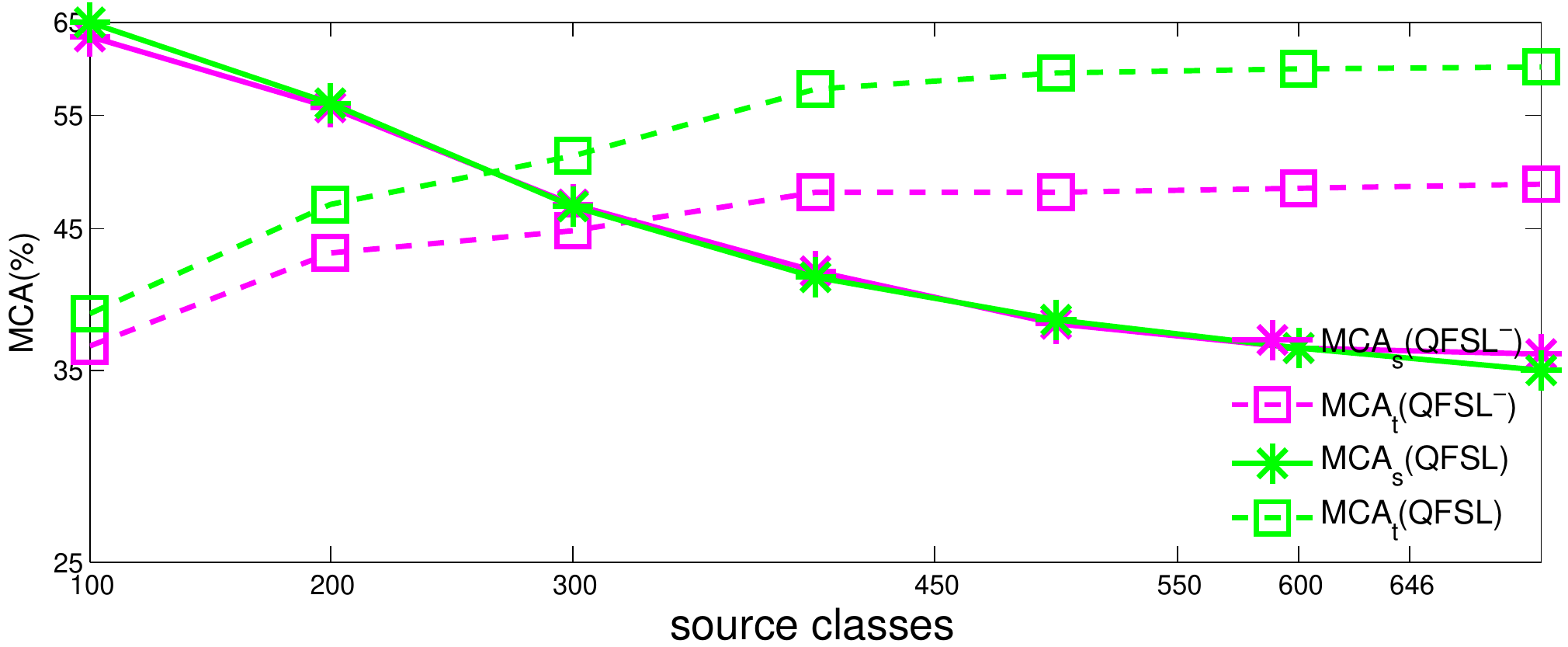}
  \caption{Performance of QFSL with different numbers of source classes on SUN. }
  \label{fig:performance_gain}
\end{figure}

Results in generalized settings are demonstrated in Figure~\ref{fig:performance_gain}. On the one hand, as the number of source classes increases, the classification task of source data becomes more difficult, which results in the performance drop in $MCA_s$. On the other hand, the increasing source classes provide more knowledge to build the mapping between the visual and the semantic embeddings, which results in the performance boost in terms of $MCA_t$.

Note that albeit with taking additional consideration of addressing the bias problem, our proposed method produces a comparable performance with the baseline QFSL$^{-}$ in $MCA_s$. Furthermore, with more imbalanced source and target classes, the new test instances from target classes are more likely to be classified into source classes (\ie, the bias problem is more severe). Because our method alleviates the bias problem, it yields much better performance in this case. Consequently, as the number of source classes increases (\ie, the imbalance ratio between source and target classes becomes larger), the superiority of our method over the baseline QFSL$^{-}$ becomes larger.

\section{Conclusions and Future Work}
\label{section:conclusion}
In this work, we have proposed a straightforward yet effective method to learn the unbiased embedding for ZSL. This method assumes both the labeled source data and the unlabeled target data are available at the training time. On the one hand, the labeled source data are projected to the points specified by the source classes in the semantic space, which builds the relationship between the visual embeddings and the semantic embeddings. On the other hand, the unlabeled target data are forced to be projected to other points specified by the target classes, which alleviates the bias towards source classes significantly. Various experiments conducted on different benchmarks demonstrate that our method outperforms other state-of-the-art ZSL methods by a large margin in both the conventional and the generalized settings.

There are many different research lines which are worthy of further study following this work. For example, in this work, semantically meaningful attributes are adopted as the semantic space. In our future work, we will exploit other semantic space such as word vectors. Another example is that this work addresses the bias problem by transductive learning, in our future work we will consider solving the same problem following the way of inductive learning.

\paragraph{Acknowledgments.} This work was supported in part by the National Basic Research Program (973 Program, No. 2015CB352400), National Natural Science Foundation of China (61572428, U1509206), National Key Research and Development Program (2016YFB1200203), Fundamental Research Funds for the Central Universities (2017FZA5014), Alibaba-Zhejiang University Joint Institute of Frontier Technologies.

{\small
\bibliographystyle{ieee}
\bibliography{egbib}
}

\end{document}